\documentclass[twoside,11pt]{article}

\usepackage{jmlr2e}

\usepackage{graphicx}
\graphicspath{{./figs/}}

\ShortHeadings{Grid Search, Random Search, Genetic Algorithm: A Big Comparison for NAS}{Liashchynskyi}
\firstpageno{1}

\begin{document}

\title{Grid Search, Random Search, Genetic Algorithm: A Big Comparison for NAS}

\author{\name Petro Liashchynskyi \email p.liashchynskyi@st.tneu.edu.ua \\
       \addr Department of Computer Engineering\\
       Ternopil National Economic University\\
       Ternopil, 46003, Ukraine
       \AND
       \name Pavlo Liashchynskyi \\
       \addr Department of Computer Engineering\\
       Ternopil National Economic University\\
       Ternopil, 46003, Ukraine}

\editor{}

\maketitle

\begin{abstract}
In this paper, we compare the three most popular algorithms for hyperparameter optimization (\textit{Grid Search, Random Search, and Genetic Algorithm}) and attempt to use them for neural architecture search (NAS). We use these algorithms for building a convolutional neural network (search architecture). Experimental results on CIFAR-10 dataset further demonstrate the performance difference between compared algorithms. The comparison results are based on the execution time of the above algorithms and accuracy of the proposed models.
\end{abstract}


\section{Introduction}
Over the last few years, convolutional neural networks (CNNs) and their varieties have seen great results on a variety of machine learning problems and applications (\cite{cnn1, Krizhevsky2012ImageNetCW, JianxinWu, NIPS2014_5485, goodfellow2014generative}). However, each of the currently known architectures is designed by human experts in machine learning (\cite{simonyan2014deep, szegedy2014going, he2015deep}). Today, the number of tasks that can be solved using neural networks is growing rapidly and designing a neural network architecture becomes a long, slow and expensive process. It's a big challenge to design a good neural network.

Typical CNN architecture consists of several convolution, pooling, and fully-connected layers. While designing a network architecture, an expert has to make a lot of design choices: the number of layers of each type (convolution, pooling, dense, etc.), the ordering of the layers, the hyperparameters for each layer, the receptive field size, stride, padding for a convolution layer, etc. 

Many kinds of research in the field of so-called automated machine learning (Auto-ML) have been made  (\cite{zhong2017yan,Path-Level}). \citet{DBZophL16} propose a reinforcement learning-based method for neural architecture search. \citet{hebbal2019bayesian} experiment with Bayesian optimization algorithm using deep Gaussian processes. \citet{Efficient} propose another approach using parameter sharing. \citet{autokeras} and \citet{adanet} presented the most inspiring frameworks for NAS and Auto-ML called \textit{Auto Keras} and \textit{AdaNet} respectively.

Despite advances in the field of automated machine learning, in this paper, we attempt to use classic hyperparameter optimization algorithms to find the optimal neural network architecture. We compare the execution time between the above algorithms and finally will know what algorithm proposes a model with the highest score for less time.

\section{Search space}
\label{sec:space}
The search space defines which neural network architectures might be discovered by used algorithm. There are many methods and strategies for neural architecture search of CNNs. In most cases, experts build architecture from scratch by alternating convolutional and fully-connected layers. 

\begin{figure}
	\centering
	\includegraphics[scale=0.8]{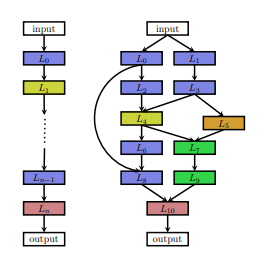}
	\caption{An illustration of different architecture spaces. Each node in the graphs corresponds to a layer in a neural network: a convolutional or pooling layer, etc. An edge from layer ${L_i}$ to
		layer ${L_j}$ denotes that ${L_j}$ receives the output of ${L_i}$ as input. Left: an element of a
		chain-structured space. Right: an element of a more complex search space with
		additional layer types and multiple branches and skip connections (\cite{elsken2018neural}.)}
	\label{fig:chain}
\end{figure}

The simple search space is the space of chain-structured neural networks, as illustrated in Figure \ref{fig:chain} (\cite{elsken2018neural}). 

Rather than designing the entire convolutional network,
one can design smaller modules and then connect them together to form a network (\cite{Efficient}). Using this approach, a neural network architecture can be written as a sequence of layers. Then the search space is parametrized by: 
\begin{itemize}
	\item $n$ number of layers;
	\item type of every layer (e.g., convolutional, pooling, fully-connected layers);
	\item hyperparameters of every layer (e.g., kernel size and filters for a convolutional layer). 
\end{itemize}

\begin{figure}
	\centering
	\includegraphics[scale=0.8]{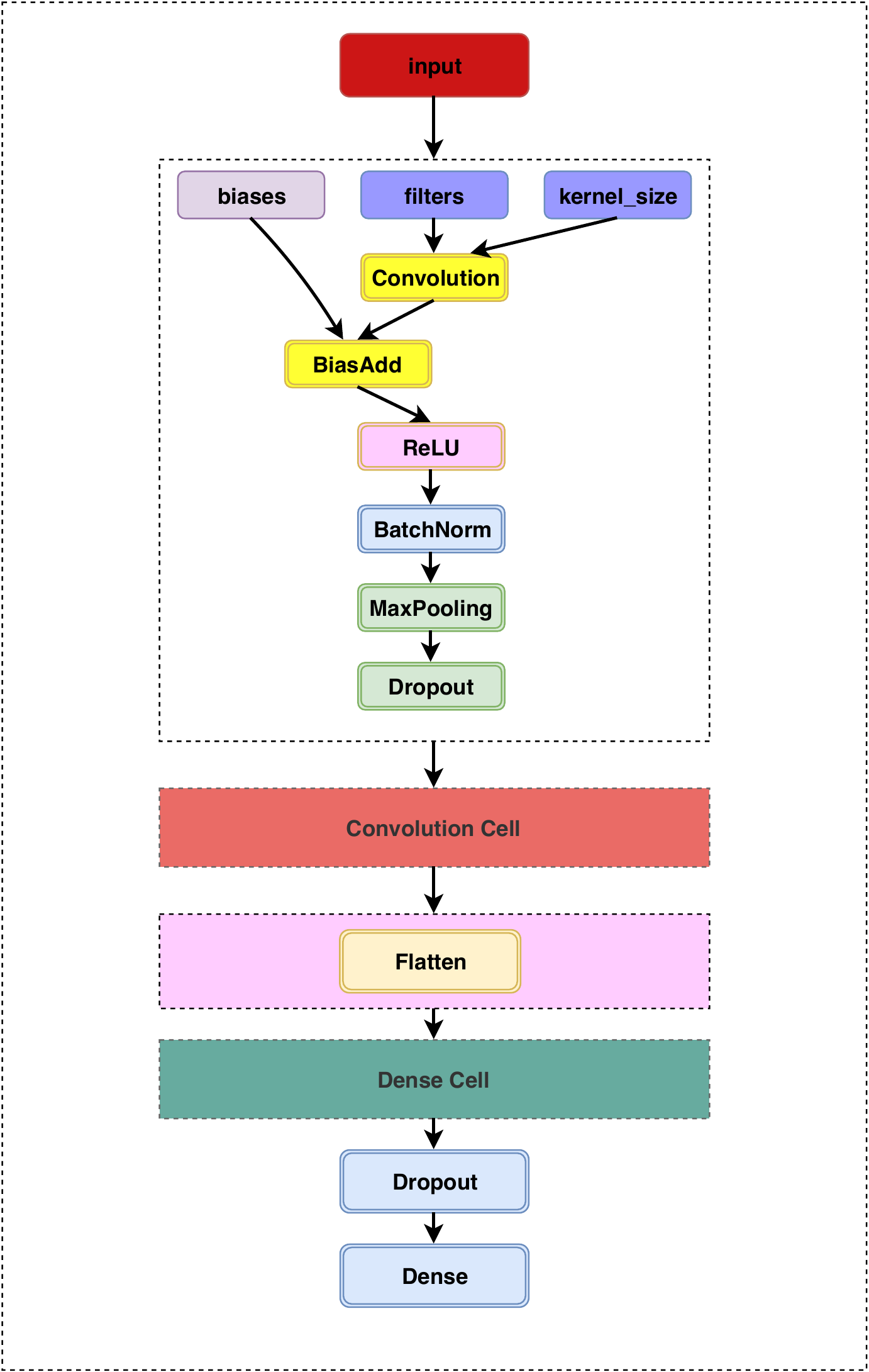}
	\caption{An illustration of a basic CNN architecture. The search space defines the number of convolutional and dense blocks.}
	\label{fig:basic}
\end{figure}

In this paper, we use the next approach: 

1) Define our basic architecture, as illustrated in Figure \ref{fig:basic}.

\begin{figure}
	\centering
	\includegraphics[scale=0.8]{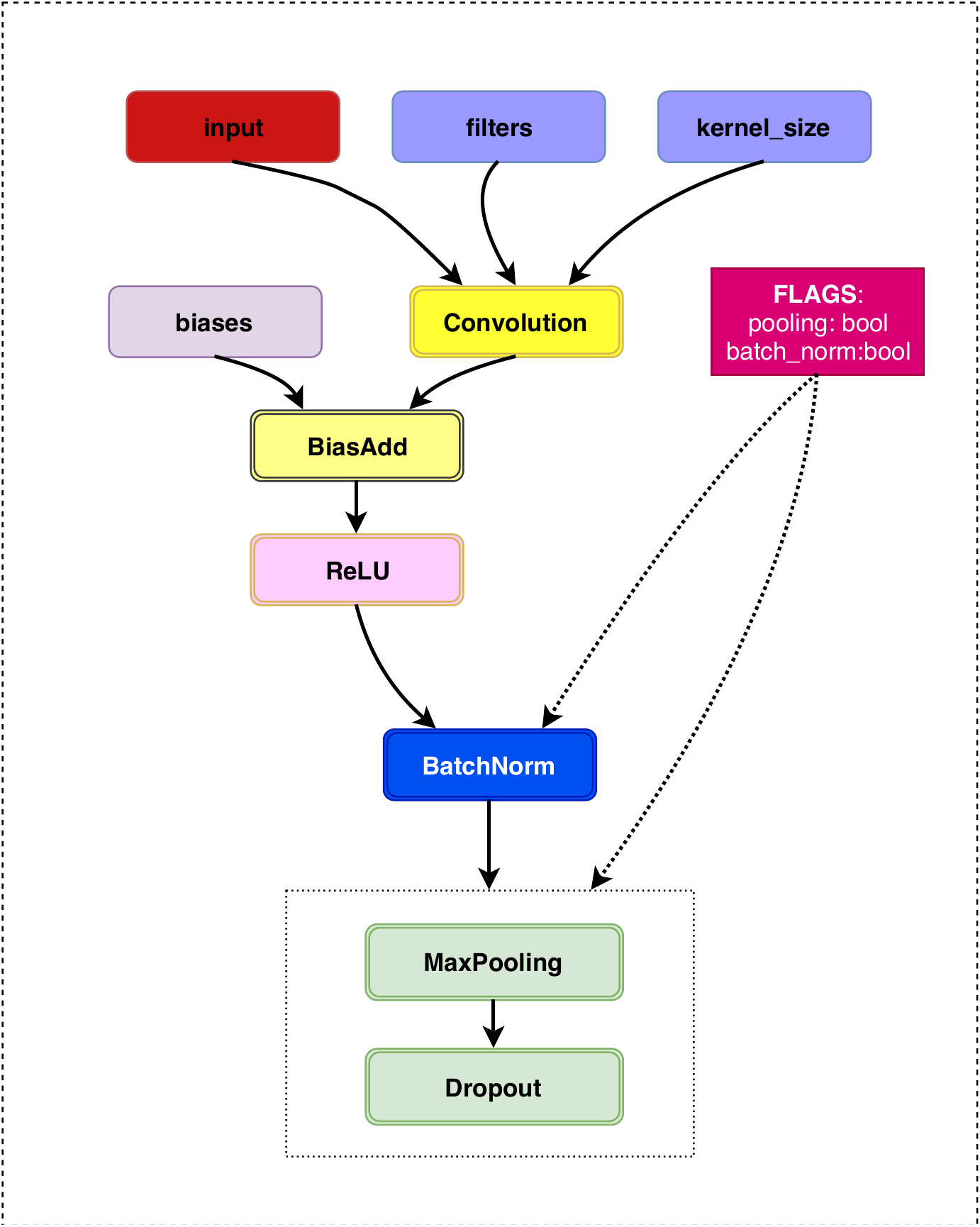}
	\caption{An illustration of a convolutional block. This block accepts the following parameters as required: \textit{input, filters, kernel size}. If additional parameters are specified, the output is passed through the \textit{BatchNormalization} and \textit{MaxPooling} layers.}
	\label{fig:fig1}
\end{figure}

2) Build a new architecture by adding the convolutional and dense cells (Figure \ref{fig:fig1}) to the basic architecture. A \textit{dense cell} is just a fully-connected layer with ReLU activation.

The choice of the search space determines the difficulty of the optimization
problem: even for the case of the search space based on a single cell with fixed architecture, the optimization problem remains difficult (\cite{elsken2018neural}) and relatively high-dimensional (since more complex models tend to perform better, resulting in more design
choices).

\section{Search strategy}

\begin{figure}
	\centering
	\includegraphics[scale=1]{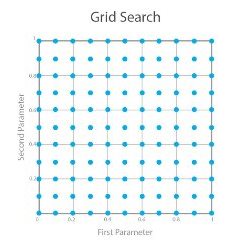}
	\caption{An illustration of a grid search space. We manually set a range of the possible parameters and the algorithm makes a complete search over them. In other words, the grid search algorithm is a complete brute-force and takes a too long time to execute.}
	\label{fig:gridy}
\end{figure}

\paragraph{Grid Search.} The traditional method of hyperparameters optimization is a grid search, which simply makes a complete search over a given subset of the hyperparameters space of the training algorithm (Figure \ref{fig:gridy}). Because the machine learning algorithm parameter space may include spaces with real or unlimited values for some parameters, it is possible that we need to specify a boundary to apply a grid search. Grid search suffers from high dimensional spaces, but often can easily be parallelized, since the hyperparameter values that the algorithm works with are usually independent of each other. 

\paragraph{Random Search.}
It overrides the complete selection of all combinations by their random selection. This can be easily applied to discrete cases, but the method can be generalized to continuous and mixed spaces. Random search can outperform a grid search, especially if only a small number of hyperparameters affect the performance of the machine learning algorithm.
\begin{figure}
	\centering
	\includegraphics[scale=1]{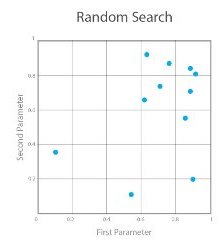}
	\caption{An illustration of a random search space. We manually set a range of bounds of the possible parameters and the algorithm makes a search over them for the number of iterations we set.}
	\label{fig:rand}
\end{figure}

\paragraph{Genetic Algorithm.}
The genetic algorithm is an evolutionary search algorithm used to solve optimization and modeling problems by sequentially selecting, combining, and varying parameters using mechanisms that resemble biological evolution. Genetic algorithms simulate the process of natural selection which means those species who can adapt to changes in their environment can survive and reproduce and go to the next generation. 

Each generation consist of a population of individuals and each individual represents a point in search space and possible solution. Each individual is represented as a string of character/integer/float/bits. This string is analogous to the chromosome.

The genetic algorithm begins with a randomly generated population
of chromosomes. Then, it makes a selection process and
recombination, based on each chromosome fitness (score). Parent
genetic materials are recombined to generate child
chromosomes producing the next generation. This process is
iterated until some stopping criterion is reached (\cite{Loussaief2018}).

\section{Experiments and Results}
\paragraph{Dataset.}  In our experiments we use the CIFAR-10 dataset with data preprocessing and augmentation procedures. This dataset consists of 50,000 training images and 10,000 test images. We first preprocess the data by mean and standart deviation normalization. Additionally, we shift the training images horizontally and vertically, and randomly flipping them horizontally. 

\paragraph{Search spaces.} We apply each of the algorithm to one search space: 
the macro search space over basic convolutional model with adding convolutional cells to it (Section \ref{sec:space}).

\paragraph{Training details.} All convolutional kernels are initialized with He uniform initialization (\citet{he2015delving}). We also apply $L_2$ weight decay with rate $10^{-4}$ and set the kernel size in convolutions to 3. The first convolutional layer uses 32 filters. Each convolutional cell uses 64 filters. All convolutions are followed by BatchNormalization with MaxPooling. The parameters of each network are trained with Adamax optimizer (\citet{kingma2014adam}), where the learning rate is set to $2e-3$ and other parameters is stay by default. We set the dropout rate to 0.2 in each convolutional cell. Dropout rate in the basic model is 0.5 and the number of units in the dense cells is 512. Each architecture search is run for 50 epochs on Nvidia Tesla K80 GPU. The basic model achieves 76\% accuracy after 50 epochs.

\subsection{Grid Search}
The possible number of the convolutional cells is set to (0, 2, 3, 4) and number of the dense cells is set to (1, 2). 

\paragraph{Results.} The whole trainig procedure of $2\times4=8$ (length of conv cells list $\times$ length of dence cells list) models took $\approx$4.3 hours.

\begin{table}[htb]
	\centering
	\def\arraystretch{1.7}
	\begin{tabular}{lllll}
		\multicolumn{3}{c|}{Model params}                & \multicolumn{2}{l}{Evaluating}  \\ 
		\hline
		Conv cells & Dense cells & \multicolumn{1}{l|}{Size} & Accuracy \% & Score\\ 
		\hline
		0           & 1                                 & 4.2M & 75$\pm$0.3 & 0.72\\
		0           & 2                                 & 4.4M & 77$\pm$0.2  & 0.74\\
		2           & 1                                 & 0.58M & 82 & 0.57\\ 
		2           & 2                                 & 0.84M & 83$\pm$0.4 & 0.57\\ 
		3           & 1                                 & 0.23M & 81.6 & 0.6\\ 
		3           & 2                                 & 0.49M & 81.8 & 0.61\\ 
		4           & 1                                 & 0.16M & 80.9 & 0.65\\ 
		4           & 2                                 & 0.43M & 80.1 & 0.66\\ 
	\end{tabular}
	\caption{Results of Grid Search}
\end{table}

As you can see, the best accuracy is about 83\%. The model has 2 convolutional and 2 dense cells. 

\subsection{Random Search} We restricted our search space by a random integer between 2 and 8 for the possible number of convolutional cells and a random integer between 1 and 4 for possible dense cells \footnote{We had also to limit the number of MaxPooling layers in the convolutional cells to prevent dimensions error when input dimension comes too low. We skip MaxPooling if the dimension of the input comes lower than (?, 2, 2, ?).}. We took 5 runs and got the following results.

\paragraph{Results.} The whole training procedure of 5 runs took $\approx$2.7 hours.

\begin{table}[htb]
	\centering
	\def\arraystretch{1.7}
	\begin{tabular}{lllll}
		\multicolumn{3}{c|}{Model params}                & \multicolumn{2}{l}{Evaluating}  \\ 
		\hline
		Conv cells & Dense cells & \multicolumn{1}{l|}{Size} & Accuracy \% & Score\\ 
		\hline
		3           & 2                                 & 0.49M & 80$\pm$0.05 & 0.67\\
		2           & 2                                 & 2.4M & 83.65$\pm$0.02  & 0.60\\
		4           & 1                                 & 0.66M & 85.8 & 0.51\\ 
		7           & 2                                 & 0.64M & 83.8 & 0.61\\ 
		3           & 1                                 & 0.62M & 85.4 & 0.49\\ 
	\end{tabular}
	\caption{Results of Random Search}
\end{table}

The best model shows about 86\% accuracy. As you can see, this algorithm is faster than Grid Search, but if we take more runs it will be much longer.

\subsection{Genetic Algorithm}
In this experiment, we set population size to 2, the number of generations to 8, and the genome length to 8. The genome of each individual has represented as a random variates Bernoulli distribution with a random state of 0.5. The first 4 bits represent the number of convolutional blocks and the rest are for the number of dense blocks \footnote{We set MaxPooling layers randomly in convolutional blocks to prevent low dimensionality of the output.}.
\paragraph{Results.} The evolutionary algorithm took about $\approx$4.13 hours to run.

\begin{table}[htb]
	\centering
	\def\arraystretch{1.7}
	\begin{tabular}{lllll}
		\multicolumn{3}{c|}{Model params}                & \multicolumn{2}{l}{Evaluating}  \\ 
		\hline
		Conv cells & Dense cells & \multicolumn{1}{l|}{Size} & Accuracy \% & Score\\ 
		\hline
		10           & 1                                 & 0.49M & 85.7 & 0.62\\
		4           & 3                                 & 2.75M & 83.6 & 0.72\\
		6           & 1                                 & 0.73M & 83.5 & 0.6\\
		
	\end{tabular}
	\caption{Results of Genetic Algorithm | Top 3 models}
\end{table}

The best model shows about 86\% accuracy. It's almost the same as other algorithms show.

\section{Final thoughts}
We tested the neural architecture search approach with the three most popular algorithms | Grid Search, Random Search, and Genetic Algorithm. Almost all of the tested algorithms take a long time to search for the best model. Grid Search is too slow, Random Search is limited to search space distributions. The most inspiring is the evolutionary algorithm, where we encode the number of parameters as a genome. Evolution may take too long before we get the best model.

Regarding hyperparameter optimization, it's difficult to say which of the above algorithms will show the best results. If a model and search space is not too large then the grid search or random search may be a good choice to do that. But if a model has too many layers and large search space, then the evolutionary algorithm may be the best choice.

\subsection{Summary} Grid search is a brute force algorithm. This makes a complete search for a given subset of the hyperparameter space. If the search space is too large, do not choose this algorithm. That is, we will train and test every possible combination of network parameters we provided.

The random search maybe a little faster, but it does not guarantee us the best results. 

Finally, if our search space is large then the best choice is the evolutionary algorithm. It also takes a long time to run, but we can control it over the number of generations and length of the population. Each individual represents a solution in the search space for a given problem. Each individual is coded as a finite length vector of components. These variable components are analogous to genes. Thus a chromosome (individual) is composed of several genes (variable components). 

When there are too many parameters for optimization, the genetic algorithm performs faster than others. Choose this one and the evolution will do all for you.

\bibliography{19-954}  

\end{document}